\documentclass{article}

\usepackage{microtype}
\usepackage{graphicx}
\usepackage{subcaption}
\usepackage{booktabs} %

\usepackage{hyperref}

\usepackage[preprint]{icml2026}

\usepackage{amsmath}
\usepackage{amssymb}
\usepackage{mathtools}
\usepackage{amsthm}

\usepackage{booktabs}
\usepackage{multirow}
\usepackage[table]{xcolor} %
\usepackage{makecell}
\newcommand{\best}[1]{\textbf{#1}}

\definecolor{rowblue}{RGB}{240,248,255} %

\usepackage[capitalize,noabbrev]{cleveref}

\theoremstyle{plain}

\theoremstyle{definition}

\theoremstyle{remark}

\usepackage[textsize=tiny]{todonotes}

\icmltitlerunning{PSA: Pyramid Sparse Attention for Efficient Video Understanding and Generation}

\begin{document}

\twocolumn[
    \icmltitle{PSA: Pyramid Sparse Attention for Efficient Video Understanding and Generation}

  \icmlsetsymbol{equal}{*}

  \begin{icmlauthorlist}
    \icmlauthor{Xiaolong Li}{equal,zju}
    \icmlauthor{Youping Gu}{equal,zju}
    \icmlauthor{Xi Lin}{equal,zju}
    \icmlauthor{Weijie Wang}{zju}
    \icmlauthor{Bohan Zhuang}{zju}
  \end{icmlauthorlist}

  \icmlaffiliation{zju}{ZIP Lab, Zhejiang University}

  \icmlcorrespondingauthor{Xiaolong Li}{xiaolong.ziplab@gmail.com}
  \icmlcorrespondingauthor{Youping Gu}{youpgu71@gmail.com}
  \icmlcorrespondingauthor{Xi Lin}{erix025@outlook.com}
  \icmlcorrespondingauthor{Weijie Wang}{wangweijie@zju.edu.cn}
  \icmlcorrespondingauthor{Bohan Zhuang}{bohan.zhuang@gmail.com}

  \icmlkeywords{Sparse Attention, Video Generation, Video Understanding, Efficient Inference}

  \vskip 0.3in
]

\begin{abstract}

Attention mechanisms are the core of foundation models, but their quadratic complexity remains a critical bottleneck for scaling. 
This challenge has driven the development of efficient attention mechanisms, with sparsity emerging as the dominant paradigm.
Current methods typically retain or discard entire key–value blocks with binary masks, resulting in substantial information loss under high sparsity. To mitigate this gap, we present \textbf{Pyramid Sparse Attention (PSA)}, a versatile module applicable to both video understanding and generation tasks. 
Instead of binary masking, PSA introduces multi-level pooled KV representations, enabling finer mask granularity. Specifically, each query block dynamically allocates lower pooling levels to critical KV blocks and higher levels to less important ones, creating an informative interpolation between full retention and complete pruning. This design, analogous to fixed-point quantization and classical feature pyramid networks in computer vision, effectively mitigates information loss while preserving computational efficiency under a low compute budget. It works with a native, hardware-friendly kernel that leverages decoupled block-tile design to ensure efficient execution.
Across video understanding and generation benchmarks, PSA preserves contextual information and visual fidelity, consistently outperforming or achieving comparable performance over existing sparse attention baselines with superior efficiency–quality trade-offs. Our code and model weights are publicly available at: \url{http://ziplab.co/PSA}

\end{abstract}
    
\printAffiliationsAndNotice{\icmlEqualContribution} 

\section{Introduction}
\begin{figure}[t]
    \centering
    \includegraphics[width=1.0\linewidth]{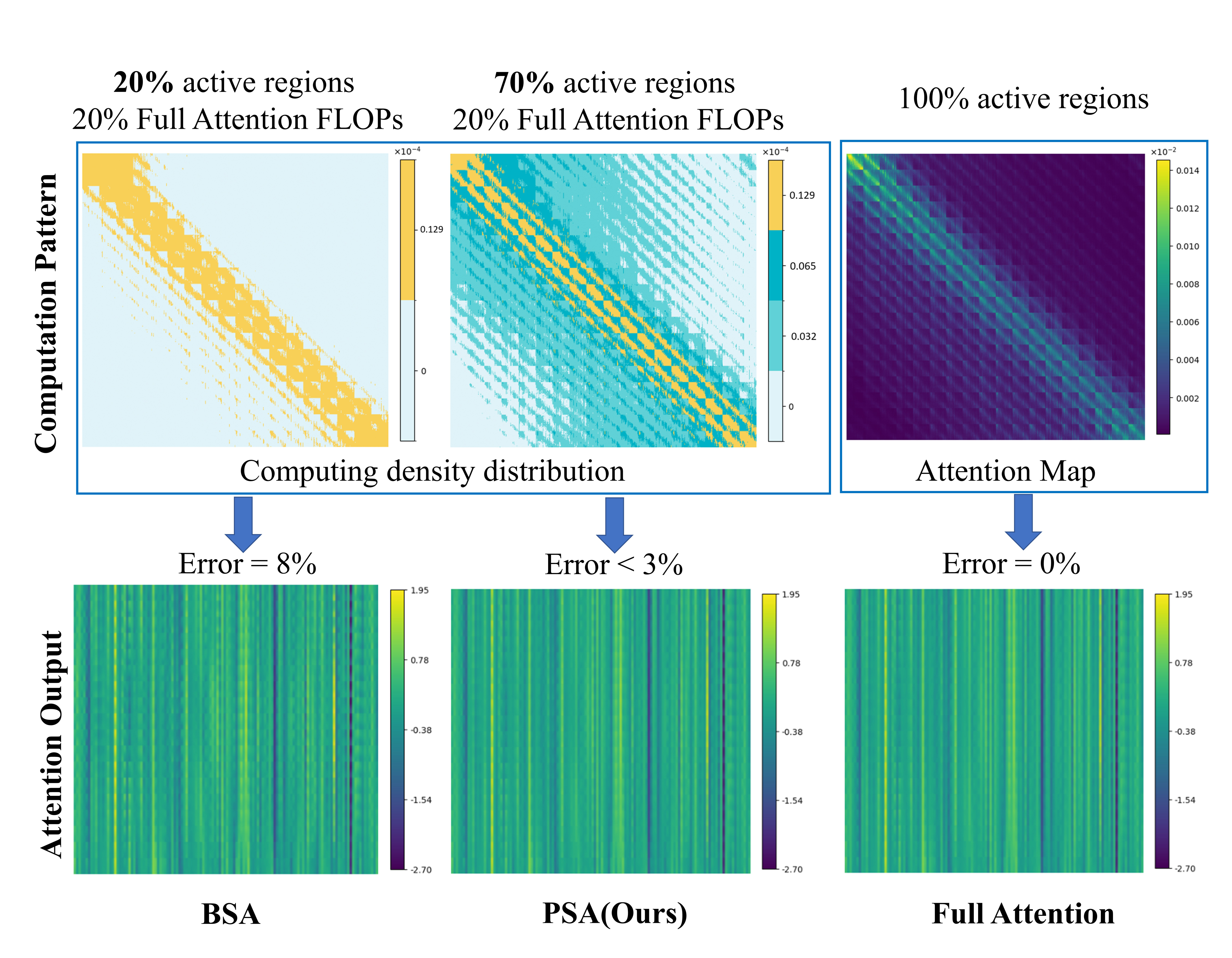}
    \caption{\textbf{Comparison of attention mechanisms under identical compute budget.} 
All methods use the same input $Q$, $K$, and $V$ tensors extracted from Wan2.1–1.3B~\cite{wan2025wanopenadvancedlargescale} denoising process.
Computation Pattern (top-left two panels):
Normalized block-wise FLOPs distribution. The two panels plot query blocks on the horizontal axis and key blocks on the vertical axis. Despite identical FLOPs ($20\%$ full), the proposed Pyramid Sparse Attention (PSA) allows each query block to attend to a much larger portion of KV blocks (70\% active regions), whereas Block Sparse Attention (BSA)~\cite{dao2022flashattentionfastmemoryefficientexact, zhang2025spargeattentionaccuratetrainingfreesparse, xu2025xattentionblocksparseattention} restricts each query to only a narrow subset of KV blocks (20\% active regions), concentrating FLOPs in limited areas.
Attention Output (bottom row):
Resulting attention visualizations. PSA closely matches the Full Attention baseline with minimal relative error ($<3\%$), while BSA shows noticeable distortions due to aggressive pruning.}

    \label{fig:psa_comparison}
    \vspace{-0.5cm}
\end{figure}
\label{sec:intro}

Recent advances in \emph{video generation} and \emph{video understanding models} have substantially increased sequence lengths, often reaching tens of thousands of tokens in modern diffusion transformers and autoregressive architectures~\cite{vaswani2023attentionneed,peebles2023scalablediffusionmodelstransformers,lin2024videollavalearningunitedvisual,wang2024qwen2vlenhancingvisionlanguagemodels,kong2025hunyuanvideosystematicframeworklarge,wan2025wanopenadvancedlargescale}. 
However, the quadratic complexity of attention has become a major obstacle to serving these models efficiently. In particular, attention computation dominates the prefill stage in long-context video understanding models and the end-to-end sampling process in video generation. For example, when processing high-resolution, long-duration videos, such as generating a $720$p clip with $81$ frames using the Wan2.1–14B~\cite{wan2025wanopenadvancedlargescale} model, inference on a single NVIDIA H20 GPU can take nearly \textit{two hours}, with attention operations alone accounting for over \textit{80\%} of the total runtime.
This prohibitive cost underscores the urgent need for more efficient attention mechanisms capable of handling long-context video inputs.

To alleviate this computational burden, recent studies exploit the inherent sparsity of the attention map 
$P=\mathrm{Softmax}\!\big(QK^\top/\sqrt{d}\big)$, 
where most elements become negligible after softmax normalization~\cite{deng2024attention}. 
Sparse attention methods leverage this property by computing only the informative regions of $P$, thereby reducing complexity without sacrificing much accuracy~\cite{zhang2025spargeattentionaccuratetrainingfreesparse,zhang2025fastvideogenerationsliding,yang2025sparsevideogen2acceleratevideo,xi2025sparsevideogenacceleratingvideo,xu2025xattentionblocksparseattention}. 
In practice, the dominant paradigm combines a \emph{mask generation strategy} with a \emph{Block Sparse Attention (BSA)} kernel that performs attention only on selected blocks for hardware efficiency~\cite{dao2022flashattentionfastmemoryefficientexact,dao2023flashattention2fasterattentionbetter,shah2024flashattention}. 
This paradigm has proven highly effective in reducing attention cost while preserving generation quality, and a series of methods following this paradigm such as XAttention~\cite{xu2025xattentionblocksparseattention} and SpargeAttention~\cite{zhang2025spargeattentionaccuratetrainingfreesparse}, have demonstrated excellent efficiency–quality trade-offs in large-scale video generation and understanding tasks.

However, existing block sparse attention mechanisms suffer from a fundamental limitation. 
At high sparsity levels, their hard binary keep/drop masks severely restrict the key–value region visible to each query, forcing the model to discard many informative areas and leading to significant information loss and performance degradation. 
Recent work such as Sparse VideoGen(SVG)~\cite{xi2025sparsevideogenacceleratingvideo,yang2025sparsevideogen2acceleratevideo}, Sliding Tile Attention~\cite{zhang2025fastvideogenerationsliding} introduce token permutation to concentrate more important key–value tokens within the limited visible block region of each query, thereby partially alleviating this issue. 
But this strategy conflicts with the design of the causal attention mask: after permutation, the causal mask becomes highly unstructured, making the algorithm difficult to implement efficiently. 
Moreover, the additional cost of reordering partially offsets the computational gains it aims to achieve.

\begin{figure}[t]
    \centering
    \includegraphics[width=\linewidth]{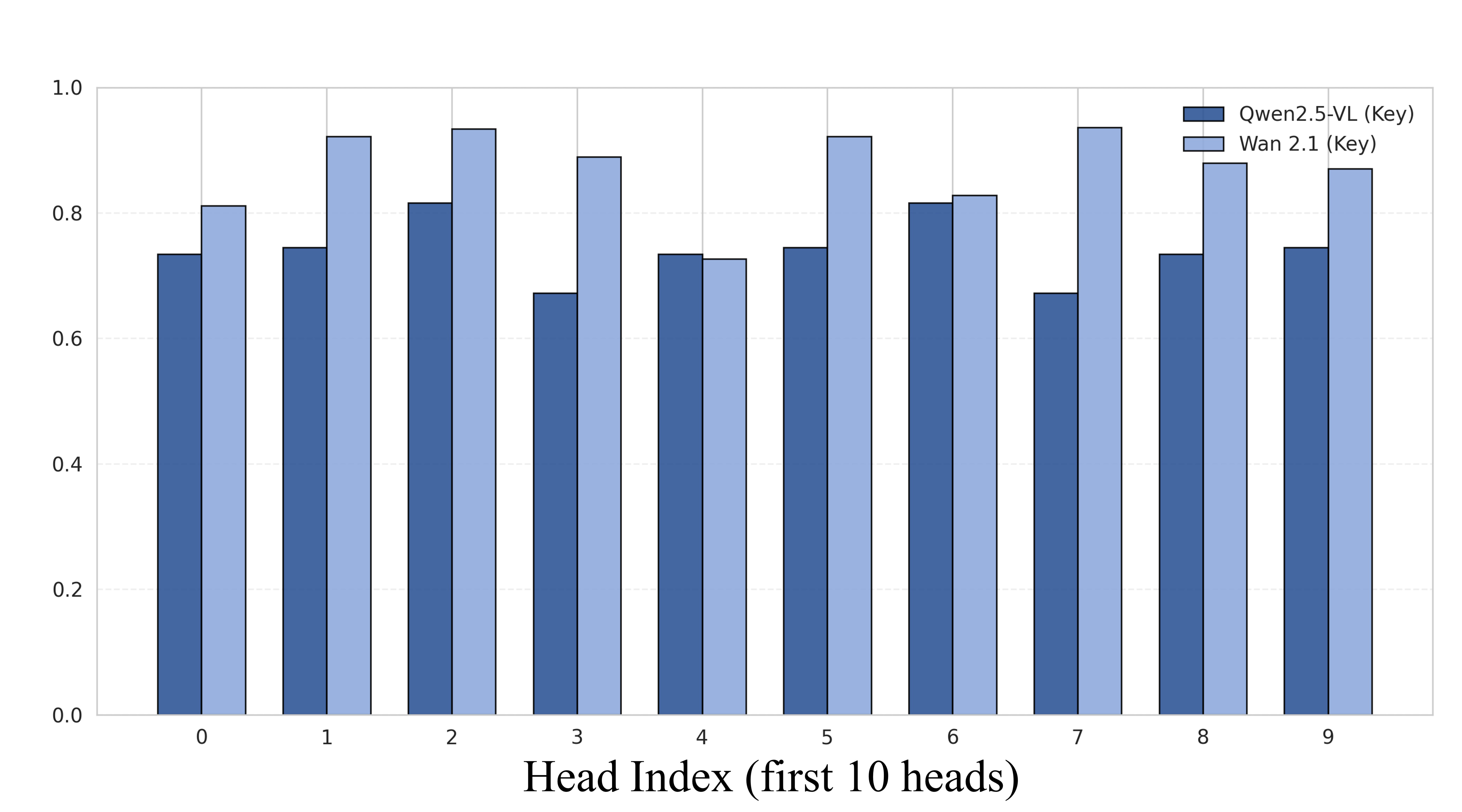}
    \caption{\textbf{Adjacent Key Token Cosine Similarity.} High cosine similarity between key tokens (Qwen2.5-VL, Wan2.1-1.3B) motivates hierarchical pooling: nearby visual tokens are highly similar.}
    \label{fig:adjacent_k_similarity}
    \vspace{-0.5cm}
\end{figure}

To this end, we propose \textit{Pyramid Sparse Attention (PSA)}, a module that preserves rich contextual information under low compute budget while remaining compatible with both causal and bi-directional attention masking.

The design of PSA is 
motivated by an empirical observation that adjacent key tokens in visual
sequences exhibit strong local similarity (~\cref{fig:adjacent_k_similarity}), suggesting that nearby keys can be aggregated by average pooling with minimal information loss. 
Moreover, a larger pooling size indicates a larger degree of information abstraction. 
Building on this insight, 
PSA replaces the binary keep/drop rule in BSA with \emph{multi-level pooled KV representations}, where important KV blocks are assigned to lower (finer) pooling levels and less important blocks to higher (coarser) levels, and the least important blocks are entirely skipped to avoid redundant computation (see ~\cref{fig:framework}), resulting in a smoother computation allocation under the same compute budget, as shown in ~\cref{fig:psa_comparison}.

An intuitive analogy can be drawn to Feature Pyramid Networks (FPNs)~\cite{DBLP:journals/corr/LinDGHHB16} in dense prediction tasks, which assign objects of different scales to distinct feature levels.
From a quantization perspective, PSA generalizes BSA's 1-bit binary mask into a multi-bit, fixed-point scheme. Here, each non-zero element indicates a specific pooling level for the KV block, while a zero value skips the block entirely, enabling finer-grained compute budget allocation.

Moreover, PSA allows each query block to allocate computation adaptively by estimating the importance of each query-key block pair based on attention scores and generating a multi-level attention mask accordingly.

As a result, as shown in ~\cref{fig:psa_comparison}, under \textit{the same compute budget}, each query in PSA attends to about \textit{70\% of the KV blocks} on average, substantially expanding its receptive field, while BSA retains only 20\% of the blocks and rigidly prunes context. Consequently, PSA produces attention outputs that are much closer to full attention, achieving a superior balance between accuracy and efficiency.

We empirically observe that PSA exhibits broad applicability across both video 
understanding and video generation tasks. On Video-MME~\cite{fu2025videommefirstevercomprehensiveevaluation} 
with Qwen2.5-VL~\cite{qwen2025qwen25technicalreport}, PSA matches the full-attention 
accuracy while achieving the minimal compute budget that preserves quality
(approximately 35\% full attention FLOPs) among all sparse-attention baselines. 
For video generation, PSA consistently outperforms prior block-sparse mechanisms in the 
training-free setting and further improves VBench~\cite{huang2023vbenchcomprehensivebenchmarksuite} 
scores when combined with the distillation framework 
TDM~\cite{luo2025learningfewstepdiffusionmodels} on CogVideoX–5B~\cite{hong2022cogvideolargescalepretrainingtexttovideo}, 
even when limited to only 15\% of the full-attention compute budget.
These results highlight PSA’s strong compatibility and plug-and-play efficiency across 
diverse architectures and tasks.

Our contributions are summarized as follows: 
(1) \emph{Multi-level retention beyond binary masks:} PSA constructs a pyramid of pooled KV blocks that expands each query’s receptive field without increasing FLOPs, thereby mitigating performance degradation under low compute budget.
(2) \emph{Hardware-friendly implementation:} 
We adopt an efficient merge–split kernel design that decouples the logical block size from the hardware tile size, making the processing of dynamically varying KV block sizes introduced by multi-level pooling more efficient.
This design maintains efficient GPU utilization, supports backpropagation, and is fully compatible with FlashAttention~\cite{dao2023flashattention2fasterattentionbetter,dao2022flashattentionfastmemoryefficientexact}.
(3) \emph{Versatile applicability:} 
Owing to our design that does not rely on token reordering, PSA can be seamlessly applied to both video understanding and generation models, consistently outperforming all Block-Sparse Attention based methods under low compute budget conditions.

\section{Related Work}

The quadratic computational and memory cost of standard attention presents a significant bottleneck for processing long sequences, particularly in tasks like video generation~\cite{kong2025hunyuanvideosystematicframeworklarge,wan2025wanopenadvancedlargescale,tan2025dsvexploitingdynamicsparsity,polyak2025moviegencastmedia,hong2022cogvideolargescalepretrainingtexttovideo} and understanding~\cite{qwen2025qwen25technicalreport,zhang2023videollamainstructiontunedaudiovisuallanguage,madan2024foundationmodelsvideounderstanding,xu2021videoclipcontrastivepretrainingzeroshot,yan2021videogptvideogenerationusing}. To overcome this limitation, sparse attention mechanisms have been developed, which apply a mask to the attention matrix to reduce computations. These methods are broadly classified as static or dynamic.

\noindent\textbf{Static sparse attention.}
Static sparsity methods employ predefined, input-agnostic attention patterns~\cite{zhang2025fastvideogenerationsliding,Hassani_2023_CVPR,child2019generatinglongsequencessparse,li2025radialattentiononlogn,xiao2024efficientstreaminglanguagemodels,zhang2023h2oheavyhitteroracleefficient,xiao2024duoattentionefficientlongcontextllm,xiao2024infllmtrainingfreelongcontextextrapolation,chen2025sparsevditunleashingpowersparse}. These include established patterns such as sliding-window attention~\cite{zhang2025fastvideogenerationsliding,Hassani_2023_CVPR,zhang2023h2oheavyhitteroracleefficient,xiao2024infllmtrainingfreelongcontextextrapolation,xiao2024duoattentionefficientlongcontextllm}, attention sink patterns~\cite{zhu2025sampleattentionnearlosslessaccelerationlong,fu2024moamixturesparseattention,xiao2024efficientstreaminglanguagemodels}, and spatiotemporal energy decay patterns~\cite{li2025radialattentiononlogn}. While computationally efficient, the primary drawback of these methods is their inherent rigidity. Because the patterns are fixed and input-agnostic, they risk missing critical long-range dependencies, which can lead to sub-optimal performance and potentially unstable generation quality.

\noindent\textbf{Dynamic sparse attention.}
To address the limitations of static patterns, dynamic sparsity methods were introduced. These methods generate an \textit{input-sensitive} mask $M$ during the forward pass, for instance, by thresholding attention scores~\cite{zhang2025vsafastervideodiffusion,gu2025bladeblocksparseattentionmeets,xu2025xattentionblocksparseattention,yang2025sparsevideogen2acceleratevideo,xi2025sparsevideogenacceleratingvideo,lu2025mobamixtureblockattention,yuan2025nativesparseattentionhardwarealigned,song2025videonsanativesparseattention}. However, while these content-aware, element-wise adaptive masks offer greater flexibility, they introduce a new challenge: hardware inefficiency. The unstructured, sparse matrix operations resulting from these fine-grained masks lead to poor memory access patterns and low hardware utilization, particularly on parallel processors like GPUs.

\noindent\textbf{Block sparse attention.}
The challenge of hardware utilization motivated the development of Block Sparse Attention (BSA)~\cite{dao2023flashattention2fasterattentionbetter,dao2022flashattentionfastmemoryefficientexact,xu2025xattentionblocksparseattention,gu2025bladeblocksparseattentionmeets,guo2024blocksparse}. This approach introduces a hardware-aware sparse structure by partitioning the $Q, K, V$ and attention matrices into coarse-grained blocks. A binary block mask $M$ then dictates whether an entire block's computation is performed or skipped. By operating at this block level, BSA preserves the dense matrix operations within each block, which is essential for achieving high throughput on modern GPUs. However, BSA still relies on a rigid, binary decision at the block level, which can lead to significant information loss under high sparsity. In contrast, our work introduces a multi-level pooled KV pyramid, allowing each query to access a substantially larger receptive field under the same computational budget and critically mitigating the information loss endemic to high-sparsity binary masking. Moreover, our design does not rely on token reordering, ensuring versatile and seamless applicability across both video understanding and generation models while maintaining high GPU throughput via an efficient kernel.

\section{Methodology}

\begin{figure}[t]
    \centering
    \includegraphics[width=\linewidth]{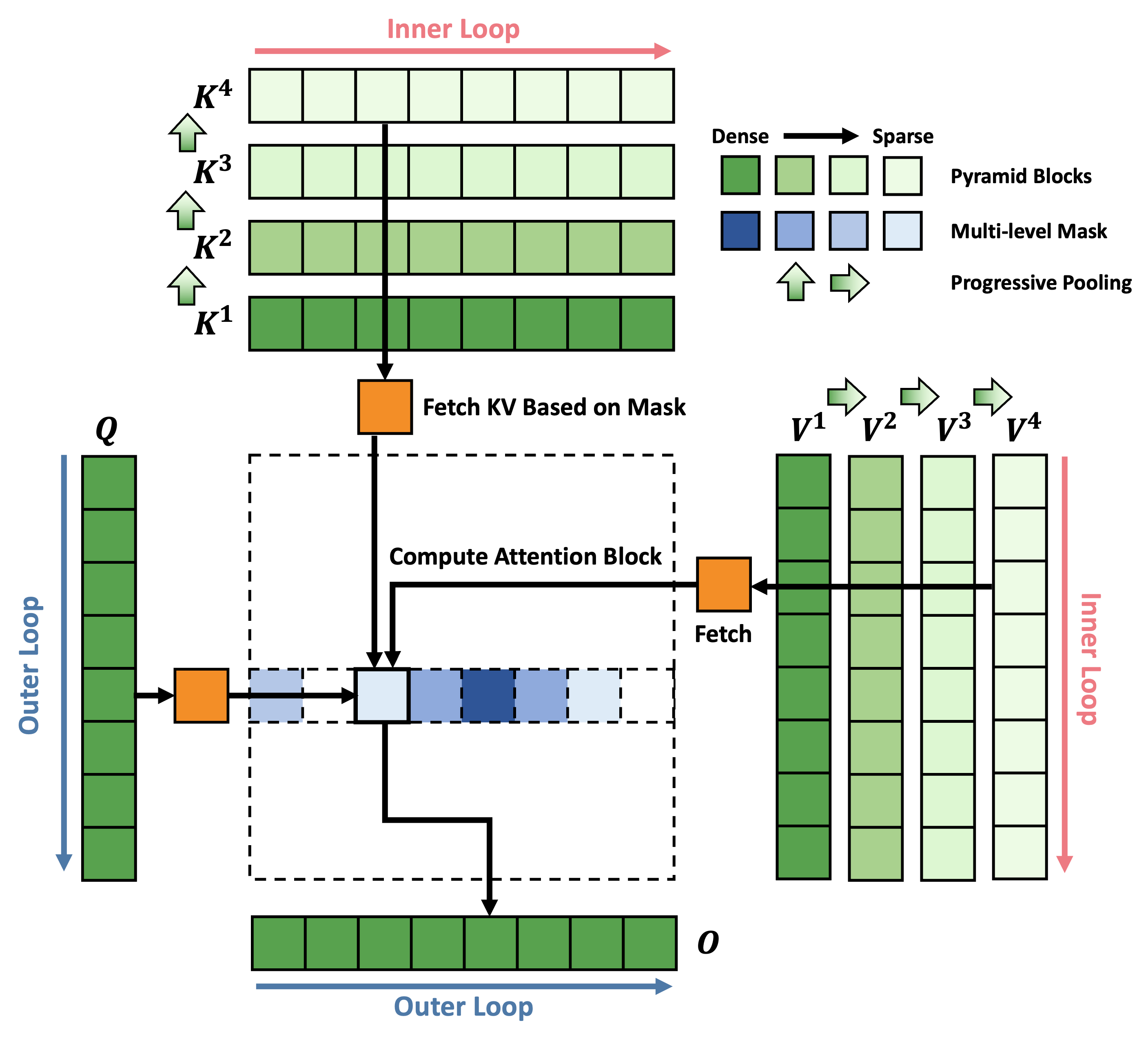}
    \caption{\textbf{Overview of the Pyramid Sparse Attention (PSA) framework.}
PSA adaptively allocates attention computation across hierarchical KV representations (green; lighter shades denote coarser levels). The multi-level mask (blue) determines which KV level each query block attends to. As illustrated, the current attention block assigned to level~4 uses the coarsest KV representation $K_j^4$ and $V_j^4$.}
\vspace{-0.5cm}
\label{fig:framework}
\end{figure}

We propose \textbf{Pyramid Sparse Attention (PSA)}, a hierarchical sparse attention framework that allocates computation adaptively across multi-level pooled key-value (KV) representations.
PSA consists of three components: (1) Pyramid KV Blocks that capture coarse-to-fine context, (2) a Multi-Level Mask Generator that predicts hierarchical sparsity, and (3) Adaptive Pyramid Attention that fetches KV blocks and computes attention accordingly (see ~\cref{fig:framework}).

\subsection{Problem Definition and Notation}

We consider the standard attention formulation with query, key, and value sequences 
$Q \in \mathbb{R}^{N \times d}$, $K, V \in \mathbb{R}^{N \times d}$,
where $N$ is  the sequence length and $d$ is the hidden dimension.
The full attention is computed as
\begin{equation}
\mathrm{Attn}(Q, K, V) = \mathrm{Softmax}\!\left(\frac{QK^\top}{\sqrt{d}}\right)V,
\end{equation}
whose quadratic complexity $O(N^2)$ becomes the bottleneck for long sequences, for example in video understanding or generation.

To reduce computation and scale attention to long sequences, block sparse attention is commonly adopted to restrict attention computation to a subset of token blocks. The query and KV sequences are divided into non-overlapping blocks of size $b_q$ and $b_k$, resulting in $n_q = N / b_q$ query blocks and $n_k = N / b_k$ KV blocks. The attention is then computed block-wise between query blocks $\{Q_i\}_{i=1}^{n_q}$ and KV blocks $\{(K_j, V_j)\}_{j=1}^{n_k}$.

A binary attention mask 
$M \in \{0,1\}^{n_q \times n_k}$
is typically used in block sparse attention, where $M_{ij}=1$ indicates that query block $Q_i$ attends to KV block $(K_j,V_j)$.
However, such binary masking enforces a hard keep-or-drop decision, which not only causes information loss from discarded regions but also limits how computation can be flexibly allocated across blocks of varying importance.

To address this limitation, we introduce a multi-level representation of KV blocks and a multi-level mask $M \in \{0, 1, \ldots, H\}^{n_q \times n_k}$, allowing a hierarchical and adaptive sparsity control.
This serves as the foundation of our proposed PSA, which dynamically allocates compute budget across coarse-to-fine levels. 
We detail the key design components in the following sections.

\subsection{Pyramid KV Blocks}

To construct multi-level KV representations, we build a hierarchical pyramid of $H$ levels by progressively pooling along the sequence dimension:
\begin{align}
K_i^{1} &= K_i, \quad &K_i^{h+1} = \mathrm{MeanPool}(K_i^{h}, 2, 2),\\
V_i^{1} &= V_i, \quad &V_i^{h+1} = \mathrm{MeanPool}(V_i^{h}, 2, 2),
\end{align}
where $\mathrm{MeanPool}(x, k, s)$ denotes 1D mean pooling with kernel size $k$ and stride $s$ along the sequence dimension.

Through this process, we obtain sets of pyramid KV blocks
$\{K_i^1, K_i^2, \dots, K_i^H\}$ and $\{V_i^1, V_i^2, \dots, V_i^H\}$,
where the $h$-th level block represents a token group of size $\frac{b_k}{2^{h-1}}$.  
As $h$ increases, the representation becomes increasingly coarser, summarizing broader contextual information within each block.
This hierarchical pooling reduces the effective KV length by a factor of $2^{h-1}$ at pyramid level $h$, enabling the model to access coarse-to-fine contextual representations and dynamically balance accuracy and efficiency under limited compute budget.

\subsection{Multi-Level Mask Generator}

Given the multi-level KV hierarchy, we estimate the importance of each query–key block pair and generate a multi-level mask to guide adaptive attention computation.

\subsubsection{Block Importance Estimation}

In this step, we compute a lightweight importance score $S_{ij}$ for each query–key block pair $(Q_i, K_j)$, which reflects how crucial it is to preserve fine-grained attention information within that region. To leverage domain-specific characteristics, we employ different importance estimators for video generation and understanding tasks.

For \textbf{video generation}, we employ a sampling-based strategy to approximate block importance efficiently. Specifically, a small subset of tokens is randomly sampled from both the query and key blocks, denoted as $\tilde{Q}_i \in \mathbb{R}^{s_q \times d}$ and $\tilde{K}_j \in \mathbb{R}^{s_k \times d}$, where $s_q \ll b_q$ and $s_k \ll b_k$. The importance score is then estimated as the maximum attention score between the sampled tokens:
\begin{equation}
S_{ij} = \max \left(\mathrm{Softmax}\!\left(\frac{\tilde{Q}_i \tilde{K}_j^\top}{\sqrt{d}}\right)\right).
\end{equation}
To improve locality and stabilize both the sampling estimation and the pyramid pooling, we apply a Hilbert curve permutation ~\cite{zhang2025spargeattentionaccuratetrainingfreesparse} to the base sequences before computation.
\label{sec:method-331}

For \textbf{video understanding}, we adopt the antidiagonal scoring ~\cite{xu2025xattentionblocksparseattention}, and introduce intra-block similarity verification to address the relatively low token similarity in video understanding (\cref{fig:adjacent_k_similarity}). This verification compares the cosine similarity of adjacent tokens against level-specific thresholds, which directly sets a maximum acceptable pooling level for the block.\label{videounderstandingtrick}

These variants demonstrate that PSA is agnostic to the specific importance estimator and readily adapts to various architectures and tasks.
\begin{algorithm}[tb]
\caption{Computation of PSA}
\label{alg:psattn}
\begin{algorithmic}[1]
\REQUIRE Query blocks $\{Q_i\}_{i=1}^{n_q}$, pyramid KV blocks $\{K_j^h, V_j^h\}_{j=1,h=1}^{n_k,H}$, mask $M$, scale factor $1/\sqrt{d}$
\ENSURE Output blocks $\{O_i\}_{i=1}^{n_q}$
\FOR{each query block $Q_i$}
    \STATE Initialize $o_i \gets 0$, $m_i \gets -\infty$, $l_i \gets 0$
    \FOR{each key block $K_j$}
        \STATE $h \gets M_{ij}$
        \IF{$h = 0$}
            \STATE \textbf{continue} \COMMENT{Skip pruned block}
        \ENDIF
        \STATE $S_{ij} \gets Q_i K_j^{h\top} / \sqrt{d} + (h-1)\ln 2$
        \STATE $m_{ij} \gets \max(\mathrm{rowmax}(S_{ij}),\, m_i)$
        \STATE $P_{ij} \gets \exp(S_{ij} - m_{ij})$
        \STATE $l_{ij} \gets l_i \exp(m_i - m_{ij}) + \mathrm{rowsum}(P_{ij})$
        \STATE $o_i \gets o_i \exp(m_i - m_{ij}) + P_{ij} V_j^h$
        \STATE $m_i \gets m_{ij}$,\quad $l_i \gets l_{ij}$
    \ENDFOR
    \STATE $O_i \gets o_i / l_i$
\ENDFOR
\STATE \textbf{return} $\{O_i\}_{i=1}^{n_q}$
\end{algorithmic}
\end{algorithm}

\subsubsection{Multi-Level Mask Assignment}

Based on the estimated importance scores $S$, we generate a multi-level mask $M$ to guide adaptive attention computation, as detailed in \cref{alg:mask}.

We introduce a \textit{multi-level mask} to assign pyramid levels dynamically.
Each entry $M_{ij} \in \{0, 1, 2, \dots, H\}$ specifies which level of pyramid KV block $(K_j, V_j)$ should be fetched when computing attention for the query block $Q_i$:
\begin{equation}
M_{ij}=h>0 \Rightarrow \text{use } K^h_j, V^h_j, \quad
M_{ij}=0 \Rightarrow \text{skip.}
\label{eq:mask}
\end{equation}
A larger $h$ indicates lower block importance and coarser KV representations, while $M_{ij}=0$ corresponds to skipping the block. This formulation allows a gradual degradation of precision instead of a binary drop, effectively mitigating abrupt information loss at high sparsity.

We leverage a threshold-based masking strategy to flexibly control the sparsity among query blocks. Specifically, we first normalize the importance scores $S$ row-wise, and then compute the cumulative scores $\hat{S}_{ij}$ by summing the descending sorted importance scores for each query block.

To translate these cumulative scores into multi-level sparsity, we define a set of thresholds as hyper-parameters that determine the pyramid level assignment. For a pyramid with $H$ levels, the thresholds are defined as:
\begin{equation}
T = {\tau_1, \tau_2, \dots, \tau_H}, \quad 0 \le \tau_1 \le \tau_2 \le \cdots \le \tau_H \le 1.
\end{equation}
These thresholds specify the score budgets allocated to each level, enabling fine-grained sparsity control. The pyramid level for each query–key block pair is assigned as:
\begin{equation}
M_{ij} =
\begin{cases}
\min\{\,t \mid \hat S_{ij} \le \tau_t\,\}, & \text{if } \hat S_{ij} \le \tau_H\\
0, & \text{if } \hat S_{ij} > \tau_H.
\end{cases}
\end{equation}

Unlike quantile-based strategies with fixed per-level quotas, our threshold-based strategy dynamically adjusts sparsity according to the cumulative importance score distribution of each query.
This adaptive mechanism allows flexible computation allocation and leads to more accurate attention estimation (see~\cref{sec:ablate_reorder_mask} for analysis).

\begin{algorithm}[tb]
\caption{Multi-Level Mask Assignment}
\label{alg:mask}
\begin{algorithmic}[1]
\REQUIRE Importance map $S \in \mathbb{R}^{n_q \times n_k}$, thresholds $T=\{\tau_1,\dots,\tau_H\}$
\ENSURE Multi-level mask $M \in \{0,\dots,H\}^{n_q \times n_k}$
\STATE $M \gets \mathrm{zeros\_like}(S)$
\FOR{$i = 1$ \textbf{to} $n_q$}
    \STATE $E_i \gets S_i / \sum_j S_{ij}$ \COMMENT{Normalize row}
    \STATE $(E_i',\, \pi_i) \gets \mathrm{sort\_desc}(E_i)$
    \STATE $\hat{E}_i \gets \mathrm{cumsum}(E_i')$
    \FOR{$j = 1$ \textbf{to} $n_k$}
        \STATE $j' \gets \pi_i(j)$ \COMMENT{Map to sorted index}
        \IF{$\hat{E}_{ij'} > \tau_H$}
            \STATE $M_{ij} \gets 0$
        \ELSE
            \STATE $M_{ij} \gets \min \{ t \mid \hat{E}_{ij'} \le \tau_t \}$
        \ENDIF
    \ENDFOR
\ENDFOR
\STATE \textbf{return} $M$
\end{algorithmic}
\end{algorithm}

\begin{figure*}[t]
    \centering
    \includegraphics[width=\textwidth]{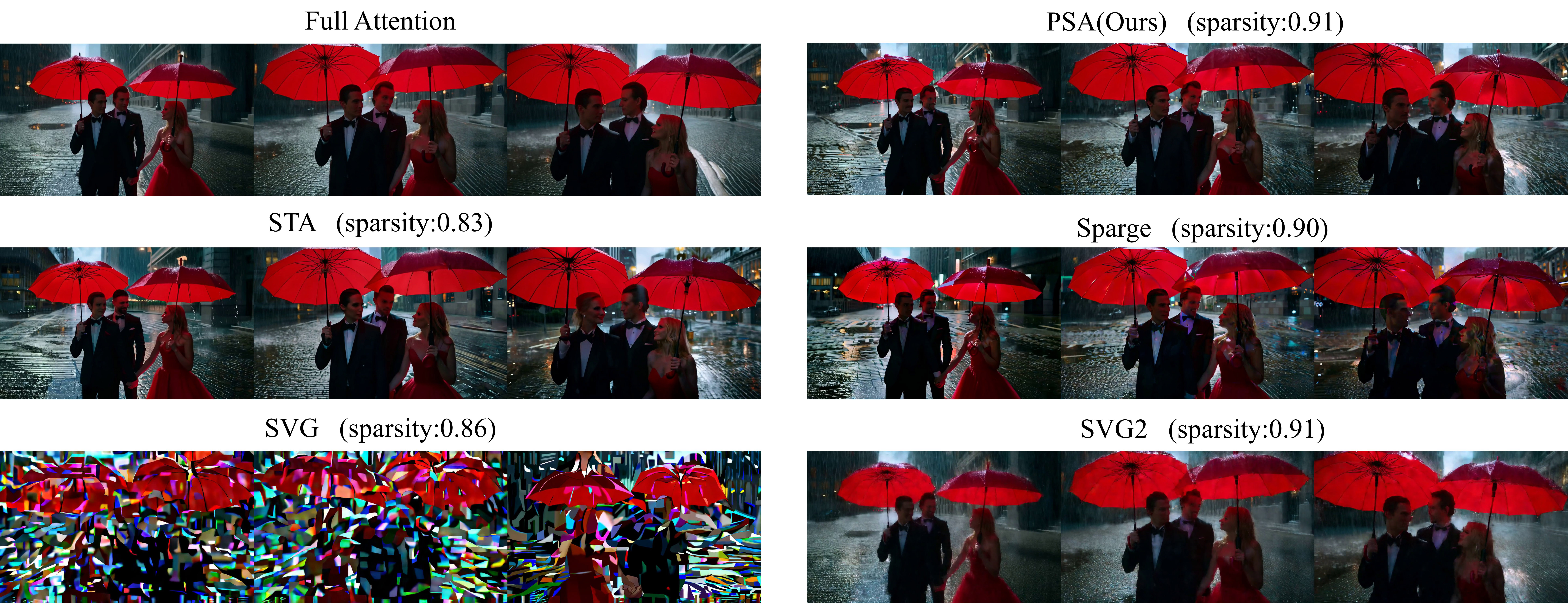}
    \caption{
    \textbf{Qualitative comparison on Wan2.1-1.3B (Text-to-Video, 720p).}
    }
    \label{fig:wan2.1_1.3b_sampled_videosl}
\end{figure*}
\subsection{Adaptive Pyramid Attention}

With the pyramid KV representations and the multi-level mask constructed, we now compute the attention in a block-wise manner. For each block, we fetch the corresponding key/value block based on the assigned pyramid level in the mask according to~\cref{eq:mask} and compute attention accordingly.

However, KV blocks in level $h$ have reduced sequence lengths due to pooling, meaning each token in $K^h_j$ represents an aggregated context of $2^{h-1}$ original tokens. To maintain consistent probability distribution in attention weights after softmax, we introduce a scaling factor of ${2^{h-1}}$ to the attention scores. This is efficiently implemented by adding a bias term $(h-1) \ln 2$ to the attention logits before softmax normalization.

Specifically, the attention score between query block $Q_i$ and key block $K_j$ at level $h$ is computed as:

\begin{equation}
A_{ij} = \mathrm{Softmax}\!\left(\frac{Q_i K_j^{h\top}}{\sqrt{d}} + (h-1)\ln 2 \right).
\end{equation}
The mechanism is detailed in~\cref{alg:psattn}.

\noindent\textbf{Complexity analysis.}
At pooling level $h$, the effective sequence length of each KV block is reduced by a factor of $2^{h-1}$, resulting in an attention cost of $O(b_q b_k / 2^{h-1})$ for that attention block.

Aggregating across all levels based on the mask distribution yields an overall expected complexity of $O(\bar{\rho} N^2)$, where $\bar{\rho}$ is the effective sparsity ratio induced by the multi-level mask:

\begin{equation}
\bar{\rho} = \frac{1}{n_q n_k} \sum_{i=1}^{n_q} \sum_{j=1}^{n_k} \rho_{ij}, \\
\text{with } \rho_{ij} = \begin{cases}\frac{1}{2^{M_{ij}-1}}, & M_{ij} > 0\\0, & M_{ij} = 0\end{cases}
\end{equation}

Compared with standard block-sparse-based attention mechanisms, PSA distributes the same compute budget more efficiently—assigning finer-grained attention to informative regions and coarser attention to redundant ones.
This adaptive allocation allows the model to retain more critical details under the same computational cost, improving representational fidelity without increasing total FLOPs.

\subsection{Hardware-Optimized Kernel Design}
\label{sec:kernel-design}

To ensure that PSA can be deployed efficiently on modern accelerators, we complement the algorithmic design with a hardware-aware implementation. The key challenge is that existing implementations often couple the logical block size with the hardware tile size, although these two hyper-parameters serve fundamentally different purposes. The block size captures the logical grouping of tokens while the tile size must be selected to maximize hardware throughput. This mismatch becomes especially pronounced in PSA because KV block pooling produces heterogeneous and often small blocks at coarser pyramid levels. Using these various block sizes directly as the execution tile leads to low compute utilization or warp divergence when the tile configuration changes dynamically.

To resolve this issue, we adopt a \textbf{decoupled block-tile design} that separates logical sequence blocks from the execution tiles used by the kernel. Building on a modified FlashAttention kernel, blocks are flexibly split or merged to match a hardware-optimized tile size. This approach allows the block size to follow the attention pattern while the tile size is tuned independently for accelerator efficiency. The design integrates seamlessly with kernel fusion, optimized memory access, and fine-grained parallelization.

The decoupled design offers broad compatibility with attention mechanisms for blocked sequences, enabling them to handle heterogeneous block structures while maintaining hardware efficiency.
In practice, our kernel preserves high tensor-core utilization even for small pooled blocks, avoids divergence across pooling levels, and achieves up to a \textbf{10$\times$} speedup over a naive PSA implementation on NVIDIA H200.

\begin{table*}[!htbp]
\centering
\footnotesize
\caption{\textbf{Quantitative comparison on Wan-series models in training-free videogen experiments.}
We report similarity metrics (PSNR, SSIM, LPIPS) and perceptual quality measures (Aesthetic Quality (Aes.), Background Consistency (Bkg.), and Imaging Quality (Img.)) from VBench~\cite{huang2023vbenchcomprehensivebenchmarksuite}.
Latency represents the average generation time per video. For clarity, the \textbf{best} result among all sparse methods under each metric is \textbf{bolded}, and the \textbf{second-best} is \underline{underlined}.}

\vspace{3pt}
\resizebox{\textwidth}{!}{
\begin{tabular}{c|c|cccccc|cc}
\toprule
\textbf{Model} & \textbf{Method} & \textbf{PSNR}$\uparrow$ & \textbf{SSIM}$\uparrow$ & \textbf{LPIPS}$\downarrow$ & \textbf{Aes.$\uparrow$} & \textbf{Bkg.$\uparrow$} & \textbf{Img.$\uparrow$}  & \textbf{Sparsity} & \textbf{Latency(s)} \\
\midrule

\multirow{6}{*}{\textbf{Wan 2.1 1.3B}} 
 & Full   & -- & -- & -- & 0.6489 & 0.9645 & 0.6557 & -- & 327 \\
  \cline{2-10}
 & SVG2  & \best{25.21} & \best{0.801} & \underline{0.126} & 0.6185 & \underline{0.9548} & 0.5545 & 0.91 & 187 \\
 & SVG   & 17.57 & 0.567 & 0.399 & 0.5039 & 0.9444 & 0.5974 & 0.85 & 165 \\
 & Sparge & 22.83 & 0.736 & 0.177 & 0.6232 & 0.9476 & 0.6409 & 0.90 & 165 \\
 & STA   & 20.56 & 0.677 & 0.197 & \underline{0.6521} & 0.9419 & \underline{0.6501} & 0.83 & 162 \\
\rowcolor{rowblue}
 & PSA(Ours)   & \underline{24.36} & \underline{0.788} & \best{0.121} &\textbf{0.6686} &\textbf{0.9612} & \textbf{0.6607}& 0.91 & 176 \\
\midrule

\multirow{5}{*}{\makecell{\textbf{Wan 2.2 5B}\\[0.2em]\small (1280 $\times$ 704, 121 frames)}}

 & Full   & -- & -- & -- & 0.6598 & 0.9564 & 0.6547 & -- & 168 \\
  \cline{2-10}
 & SVG2  & \textbf{24.25} & \textbf{0.818} & \textbf{0.092} & \underline{0.6495} & \underline{0.9518} & \underline{0.6025} & 0.90 & 149 \\
 & SVG   & 18.89 & 0.645 & 0.266 & 0.5539 & 0.9386 & 0.5877 & 0.86 & 122 \\
 & Sparge & 19.53 & 0.660 & 0.229 & 0.5482 & 0.9289 & 0.5650 & 0.89 & 124 \\
\rowcolor{rowblue}
 & PSA(Ours)   & \underline{23.03} & \underline{0.794} & \underline{0.096} & \textbf{0.6588} & \textbf{0.9569} & \textbf{0.6438} & 0.89 & 131 \\
\midrule

\multirow{6}{*}{\textbf{Wan 2.1 14B}} 
 & Full   & -- & -- & -- & 0.6918 & 0.9639 & 0.6247 & -- & 1548 \\
 \cline{2-10}
 & SVG2  & \textbf{24.79} & \textbf{0.807} & \textbf{0.085} & \underline{0.6614} & \underline{0.9439} & 0.5555 & 0.87 & 913 \\
 & SVG   & 19.84 & 0.649 & 0.300 & 0.5337 & \textbf{0.9501} & 0.5479 & 0.85 & 830 \\
 & Sparge & 22.19 & 0.737 & 0.182 & 0.6083 & 0.8779 & 0.5977 & 0.88 & 855 \\
 & STA   & 20.83 & 0.694 & 0.185 & 0.6544 & 0.9399 & \textbf{0.6489} & 0.83 & 815 \\
\rowcolor{rowblue}
 & PSA(Ours)   & \underline{23.83} & \underline{0.768} & \underline{0.105} & \textbf{0.6776} & 0.9261 & \underline{0.6400} & 0.88 & 887 \\
\bottomrule
\end{tabular}
}
\label{tab:videogen-results}
\vspace{-4pt}
\end{table*}

\label{sec:experiments}

\section{Experiments}

\subsection{Experimental Settings}

\textbf{Models.} We evaluate PSA on open-source video generation models and video understanding models: Wan2.1-\{1.3B, 14B\}~\cite{wan2025wanopenadvancedlargescale}, Wan2.2-5B~\cite{wan2025wanopenadvancedlargescale}, CogVideoX-5B~\cite{hong2022cogvideolargescalepretrainingtexttovideo}, and Qwen2.5-VL-7B~\cite{qwen2025qwen25technicalreport}.

\noindent\textbf{Baselines.}
We compare our method, Pyramid Sparse Attention (PSA), against several representative sparse attention baselines, including 
Sparse VideoGen (SVG)~\cite{xi2025sparsevideogenacceleratingvideo}, 
Sparse VideoGen2 (SVG2)~\cite{yang2025sparsevideogen2acceleratevideo}, 
Sliding-Tile Attention (STA)~\cite{zhang2025fastvideogenerationsliding}, 
SpargeAttention (Sparge)~\cite{zhang2025spargeattentionaccuratetrainingfreesparse}, and
XAttention~\cite{xu2025xattentionblocksparseattention}.
For brevity, we use abbreviated names (e.g., SVG2, STA, PSA) throughout the remainder of this section.

\noindent\textbf{Implementation details.}
All experiments are conducted on NVIDIA H200 GPUs. Unless stated, videos are generated at 1280×768 / 69 frames. 
All sparse baselines are evaluated using their official implementations without any additional optimization tricks.
In video understanding experiments, PSA reuses the same block-importance estimation based on antidiagonal scoring and incorporates an additional similarity-based constraint, as detailed in 
\cref{sec:method-331}
. All sparse attention mechanisms are applied only during the prefill stage.

\noindent\textbf{On sparsity accounting.}
PSA allocates multi-level compute to KV blocks rather than using binary keep/drop 
strategy; its reported sparsity thus denotes the sparsity of 
BSA-based method that uses the same compute budget.

\noindent\textbf{Metrics and dataset.}
For the training-free video generation experiments, we evaluate the similarity between the generated videos and their full-attention counterparts using Peak Signal-to-Noise Ratio (PSNR), Structural Similarity Index Measure (SSIM), and Learned Perceptual Image Patch Similarity (LPIPS)~\cite{zhang2018unreasonableeffectivenessdeepfeatures}. 
Beyond these similarity measures, we further adopt three perceptual dimensions from the VBench Score~\cite{huang2023vbenchcomprehensivebenchmarksuite}---widely used in recent video generation benchmarks to assess visual quality: 
Aesthetic Quality (Aes.),  
Background Consistency (Bkg.), 
and Imaging Quality (Img.).

For the distillation experiments on CogvideoX-5B~\cite{hong2022cogvideolargescalepretrainingtexttovideo}, we primarily use the VBench Score to measure the perceptual quality of videos generated by distilled models.
Our distillation process is guided by a dataset of 10,000 text prompts, sampled from the JourneyDB benchmark~\cite{sun2023journeydbbenchmarkgenerativeimage}. 
To enhance prompt quality and diversity, each sample is refined using the Qwen2.5-3B-Instruct model~\cite{qwen2025qwen25technicalreport}.

For video understanding, we adopt the Video-MME (1fps) dataset~\cite{fu2025videommefirstevercomprehensiveevaluation} to evaluate the performance of Qwen2.5-VL-7B in video understanding scenarios.

\subsection{Training-Free Video Generation}
\begin{table}[h!]
\centering
\vspace{-4pt}
\caption{Combining PSA with TDM on CogVideoX-5B.}
\label{tab:distillation}
\resizebox{\columnwidth}{!}{%
\begin{tabular}{lccc}
\toprule
\textbf{Method} & \textbf{Sparsity} & \textbf{Sampling Steps} & \textbf{VBench Score} \\
\midrule
FullAttn & -- & 50 & 0.819 \\
Distill-only & -- & \textbf{4} & 0.818 \\
\textbf{Ours} & $\mathbf{0.85}$ & \textbf{4} & \textbf{0.826} \\
\bottomrule
\end{tabular}%
} 
\vspace{-2pt}
\end{table}
\begin{table}[h!]
\centering
\vspace{-1pt}
\caption{
{Comparison of attention mechanisms on Video-MME.}
}

\label{tab:video_nderstanding}
\resizebox{\columnwidth}{!}{%
\begin{tabular}{lccccc}
\toprule
\textbf{Method} & \textbf{Short} & \textbf{Medium} & \textbf{Long} & \textbf{Overall} & \textbf{Sparsity} \\
\midrule
Full Attention & \textbf{0.752} & 0.663 & 0.537 & 0.651 & --- \\
\hline
XAttention & 0.748 & 0.661 & \textbf{0.544} & 0.651 & 0.58 \\
SpargeAttention & 0.749 & 0.663 & 0.539 & 0.650 & 0.37 \\
\hline
\textbf{PSA(Ours)} & 0.748 & \textbf{0.673} & 0.542 & \textbf{0.654} & \textbf{0.65} \\
\bottomrule
\end{tabular}%
}
\vspace{-2pt}
\end{table}
\label{traingfreevideogen}
A qualitative comparison of sparse attention mechanisms is shown in \cref{fig:wan2.1_1.3b_sampled_videosl}. All methods use the same text-to-video scene (a couple walking under a red umbrella in the rain) under similar sparsity. STA~\cite{zhang2025fastvideogenerationsliding} exhibits structure jumps and identity swaps across frames; SVG~\cite{xi2025sparsevideogenacceleratingvideo}collapses at high sparsity with random color blocks; SVG2~\cite{yang2025sparsevideogen2acceleratevideo} is more stable but overly smooth with distorted backgrounds; Sparge~\cite{zhang2025spargeattentionaccuratetrainingfreesparse} preserves layout yet shows local color distortions; Full Attention remains high-fidelity but costly. In contrast, PSA at 0.91 sparsity maintains sharp details and temporal coherence (stable contours, saturated umbrella, natural reflections), approaching the dense baseline’s visual quality with far less computation.

\cref{tab:videogen-results} further shows that under comparable high sparsity, PSA consistently surpasses SVG, STA, and Sparge on all similarity metrics (PSNR/SSIM/LPIPS) and on most perceptual axes (Aes./Bkg./Img.) across model sizes. 
Against the strongest training-free baseline SVG2, PSA delivers \emph{clearly better perceptual quality} at high sparsity---higher Aesthetic and Imaging scores on \emph{all} model sizes ---while keeping similarity metrics within small margins. 
Together with the visuals in \cref{fig:wan2.1_1.3b_sampled_videosl}, these results indicate that PSA is the \emph{most quality-preserving} sparse attention mechanism at high sparsity among all compared methods. Additional visual comparisons are provided in the appendix.

\subsection{Distillation}

To explore the generality of PSA and achieve maximum inference acceleration, we combined our method with the \textit{data-free} distillation technique TDM~\cite{luo2025learningfewstepdiffusionmodels,gu2025bladeblocksparseattentionmeets}. We integrated PSA into the student model during the distillation training phase. As shown in~\cref{tab:distillation}, this simple combination achieves a $30.2\times$ denoising time speedup over the original 50-step model, without any loss in generation quality.

Our combined approach, using $85\%$ sparsity, achieves the highest VBench score (0.826). This surpasses both the 4-step distilled baseline and even the original 50-step model. This result demonstrates that PSA is a highly compatible plug-and-play module that can be compounded with other methods like distillation to maximize inference efficiency.

\begin{table}[h!]
\centering
\caption{Comparison of multi-level vs.\ binary  masking on Wan2.1-1.3B (480p, training-free). 
The first 25\% of sampling steps do not use sparse attention.
$T=\{\tau_1,\dots,\tau_H\}$ are the cumulative  thresholds for level assignment.}
\label{tab:ablation_multilevel_mask}
\resizebox{\columnwidth}{!}{%
\begin{tabular}{lccccl}
\toprule
\textbf{Method} & \textbf{Sparsity}$\uparrow$ & \textbf{PSNR}$\uparrow$ & \textbf{SSIM}$\uparrow$ & \textbf{LPIPS}$\downarrow$ \\
\midrule
Multi-level mask & \textbf{0.79} & \textbf{23.35} & \textbf{0.856} & \textbf{0.116} \\
Binary mask & 0.75 & 23.11 & 0.851 & 0.122  \\
\bottomrule
\end{tabular}%
}
\end{table}
\begin{table}[h!]
\vspace{-4pt}
\centering
\caption{Ablation of reordering and mask-generation strategy on Wan2.1-1.3B (480p).}
\label{tab:ablate_reorder_mask}
\resizebox{\columnwidth}{!}{%
\begin{tabular}{llcccc}
\toprule
\textbf{Mode} & \textbf{Reorder} & \textbf{Sparsity}$\uparrow$ & \textbf{PSNR}$\uparrow$ & \textbf{SSIM}$\uparrow$ & \textbf{LPIPS}$\downarrow$ 
\\
\midrule
Threshold-based         & Hilbert & 0.84 & \textbf{23.78} & \textbf{0.811} & \textbf{0.085} \\
Quantile-based                & Hilbert & 0.85 & 22.54 & 0.775 & 0.118 \\
Quantile-based & None    & 0.85 & 21.81 & 0.707 & 0.201 \\
\bottomrule
\end{tabular}%
}
\vspace{-8pt}
\end{table}
\subsection{Training-Free Video Understanding}

On Qwen2.5-VL-7B, PSA delivers the best overall performance on Video-MME~\cite{fu2025videommefirstevercomprehensiveevaluation} under 
\textbf{the highest sparsity level (0.65)}. As shown in \cref{tab:video_nderstanding}, it matches or exceeds the full baseline 
across most categories---notably improving the \textit{Medium} and \textit{Long-Video} 
scores---while remaining competitive on \textit{Short Video}. These results show that 
PSA preserves (and in some cases improves) video-understanding quality at 
substantially higher sparsity, demonstrating the strongest quality-retaining 
sparsity among all compared methods in the training-free setting.

\subsection{Ablation Study}

\noindent 

\noindent\textbf{Effectiveness of multi-level mask vs.\ binary mask.} In terms of configuration, the multi-level mask group is set to $T=\{0.70,\,0.80,\,0.90,\,0.90\}$, 
while binary mask uses level ratios $T=\{0.85,\,0.85,\,0.85,\,0.85\}$, which means that only dense KV blocks are kept and no pooled representation is used.

On Wan2.1-1.3B (480p, training-free), we compare our multi-level masking  with a conventional 0/1 binary mask. The first 25\% of sampling steps do not use sparse attention.
As shown in ~\cref{tab:ablation_multilevel_mask}, PSA yields improved PSNR and SSIM, and reduced LPIPS, despite operating under a higher sparsity setting than the baselines.

\noindent\textbf{Reordering \& mask strategy (Videogen).} \label{sec:ablate_reorder_mask}
We evaluate different reordering and mask-generation strategies on Wan2.1-1.3B (480p, training-free setting) and the first 25\% of sampling steps do not use sparse attention.

We compare two multi-level masking rules (threshold-based  vs.\ quantile-based) and the role of token reordering. 
The quantile-based rule processes each query row by sorting 
block importance $S_{i,:}$ and assigning 
pooling degrees by fixed percentile cut points $0<a<b<c\le d\le 1$ (e.g., $[0,a)\!\to\!1$, $[a,b)\!\to\!2$, $[b,c)\!\to\!3$, $[c,d)\!\to\!4$, $(d,1]\!\to\!0$). 
For reordering, we follow SpargeAttention and apply a HilbertCurve permutation to  $Q, K, V $~\cite{zhang2025spargeattentionaccuratetrainingfreesparse}. Hilbert denotes a Hilbert-curve–based permutation, while None indicates no reordering.

As shown in ~\cref{tab:ablate_reorder_mask}, under near-matched sparsity, 
Threshold-based + reordering consistently outperforms quantile-based + reordering in reconstruction quality. 
Removing the reordering step from the quantile-based variant further degrades results. 
As observed in SpargeAttention~\cite{zhang2025spargeattentionaccuratetrainingfreesparse}, 
this reordering operation enhances the similarity between adjacent tokens. 
When incorporated into PSA, this property naturally complements the hierarchical pooling mechanism, 
as higher local token similarity leads to smoother multi-level aggregation and more stable reconstruction under high sparsity. 
In contrast to the quantile-based rule, the threshold-based strategy dynamically adjusts the proportion of each mask based on the input distribution, 
allowing PSA to better adapt to diverse sparsity patterns and preserve visual fidelity.

\noindent Additional ablations on the similarity constraint and variable block sizes, which also influence the performance of PSA, are presented in the supplementary material 

\section{Conclusion}
We propose Pyramid Sparse Attention (PSA), a versatile sparse attention mechanism validated across both video understanding and video generation tasks. Our method introduces multi-level sparsity beyond binary keep/skip choices, where higher sparsity levels use larger pooling degrees of KV representations; in this way, unlike block-sparse attention that fully drops low-score blocks, we preserve substantially more information under the same compute budget. As a result, PSA demonstrates strong generalization and consistently outperforms all existing Block Sparse Attention based methods under low compute budget, highlighting its robustness across diverse  model architectures and application scenarios.

\bibliography{new}
\bibliographystyle{icml2026}
\newpage
\appendix
\onecolumn
\renewcommand{\thesection}{\Alph{section}}
\setcounter{section}{0}
\newpage
\appendix
\onecolumn

\renewcommand{\thesection}{\Alph{section}}
\setcounter{section}{0}

\section{Appendix}

Due to space limitations in the main paper, we provide additional
technical details, ablation studies, and qualitative comparisons in
this appendix. The material is organized as follows:

\begin{itemize}
    \item \textbf{Sec.~B: Implementation Details of PSA.}\\
    Detailed description of PSA modules, including the cosine similarity-based
    pooling constraint, block size variants, and importance pooling operators.

    \item \textbf{Sec.~C: Additional Ablations.}\\
    Comprehensive ablation experiments on cosine similarity, block size,
    pooling operators, and multi-level allocation strategies across video
    understanding and video generation tasks.

    \item \textbf{Sec.~D: Additional Visual Comparisons.}\\
    Extended qualitative examples for the training-free video generation
    evaluation, following the same layout as the main paper.
\end{itemize}

The following sections provide the full details and results.

\section{Implementation Details of PSA}
\label{sec:supp-impl}

\subsection{Cosine Similarity Based Pooling Constraint}
\label{sec:supp-sim}

On top of the importance driven mask $M$, PSA optionally incorporates a
cosine similarity based pooling constraint.
This module is inspired by the cosine similarity constraint design in SpargeAttention~\cite{zhang2025spargeattentionaccuratetrainingfreesparse},
but we generalize it to our multi-level pooling regime.
The intuition is that key blocks whose tokens are internally similar can be safely assigned
to coarser pyramid levels, whereas blocks with heterogeneous tokens should remain at finer levels.

For PSA with $H$ levels, we introduce $H-1$ cosine similarity thresholds
\[
T_s=\{\tau_s^{(2)},\tau_s^{(3)},\dots,\tau_s^{(H)}\},
\]
where each $\tau_s^{(h)}$ specifies the minimum intra-block cosine similarity required for a block to be eligible for level~$h$. Given these thresholds, we compute the per-block maximum admissible level $L \in \{1,\dots,H\}^{n_k}$ using \cref{alg:sim-constraint}, which evaluates intra-block similarities at strides corresponding to different pooling sizes. A block may enter level~$h$ only if its stride-$2^{h-1}$ similarity exceeds $\tau_s^{(h)}$. The final mask used by PSA is obtained by combining this constraint with the importance-driven mask $M$ via
\[
\tilde{M}_{ij} = \min\bigl(M_{ij},\,L_j\bigr),
\]
ensuring that no block is assigned a level coarser than what its similarity permits.

In our experiments, we consider a pyramid with $H=4$ levels and use different similarity thresholds $(\tau_2,\tau_3,\tau_4)$ depending on the task. For Video-MME, we adopt thresholds of $(0.75,\,0.70,\,0.70)$, while for the video generation preset \textit{PSA (sim)} in Table~\ref{tab:videogen-sim-ablation}, we use $(0.70,\,0.65,\,0.60)$. The \textit{no-sim} variant is obtained by setting all thresholds to $-1$, which yields $L_j \equiv H$ and disables the similarity constraint entirely. Conversely, the \textit{1-level} variant sets all thresholds to $1$, enforcing $L_j = 1$ for all blocks and collapsing the multi-level structure into a single fine-grained level.

\begin{algorithm}[t]
\caption{Cosine Similarity-Based Pooling Level Constraint}
\label{alg:sim-constraint}
\begin{algorithmic}
  \STATE {\bfseries Input:} key blocks $\{K_j\}_{j=1}^{n_k}$, thresholds $\tau_s^{(2)}, \dots, \tau_s^{(H)}$
  \STATE {\bfseries Output:} maximum admissible levels $L \in \{1,\dots,H\}^{n_k}$
  \STATE Initialize $L_j \gets 1$ for all $j$
  \FOR{$j = 1$ {\bfseries to} $n_k$}
    \STATE Interpret $K_j$ as a tensor in $\mathbb{R}^{B \times n_{\text{head}} \times b_k \times d}$
    \FOR{$h = 2$ {\bfseries to} $H$}
      \STATE Compute $\text{sim}_h(j)$ as the mean cosine similarity of stride-$2^{h-1}$ pairs in $K_j$
      \IF{$\text{sim}_h(j) > \tau_s^{(h)}$}
        \STATE $L_j \gets \max(L_j,\,h)$
      \ENDIF
    \ENDFOR
  \ENDFOR
  \STATE \textbf{return} $L$
\end{algorithmic}
\end{algorithm}

\subsection{Block Size and Pooling Variants}

The query and key block sizes determine the granularity at which importance scores and masks are computed, and are conceptually independent of the hardware tile sizes used within the kernel implementation. Owing to the decoupled block-tile design introduced in \cref{sec:kernel-design}, PSA can employ moderate block sizes $(b_q, b_k)$ (e.g., $(32,32)$, $(64,64)$, $(128,64)$) while still achieving relatively high tensor-core utilization.

To examine the role of the pooling operator in the importance definition (\cref{sec:method-331}), we additionally evaluate a \emph{mean-pooling} variant that replaces the max operator with an average over the sampled attention scores, while keeping all other components (sampling, permutation, mask generation, and similarity constraint) identical. The corresponding ablation results are shown in Table~\ref{tab:blocksize-pooling-ablation}. 

Unless otherwise noted, all video generation experiments in this appendix are performed on Wan2.1-1.3B at 480p under the same training-free setup as in the main paper, with sparse attention enabled after the first $25\%$ of sampling steps and sparsity reported as FLOP-equivalent sparsity relative to full attention.

\section{Additional Ablations}
\label{sec:supp-ablation}

\subsection{Cosine Similarity (Video Understanding)}

We first evaluate the cosine similarity-based pooling constraint in the
video understanding setting.
Table~\ref{tab:vla-sim-ablation} compares a dense FlashAttention2 baseline with two PSA variants on Video-MME, where all sparse variants are matched at
approximately $0.65$ FLOP-equivalent sparsity.

\begin{table}[t]
\centering
\small
\setlength{\tabcolsep}{3pt}
\caption{Effect of the cosine similarity constraint on Video-MME.
All sparse variants are matched at $\sim 0.65$ FLOP-equivalent sparsity.}
\label{tab:vla-sim-ablation}
\begin{tabular}{lcccc}
\toprule
Method      & Short & Medium & Long  & Overall \\
\midrule
FA2 (dense) & 0.752 & 0.663  & 0.537 & 0.651 \\
\midrule
PSA w/o sim & 0.747 & 0.667  & 0.529 & 0.647 \\
PSA w/ sim  & \textbf{0.748} & \textbf{0.673}  & \textbf{0.542 }& \textbf{0.654} \\
\bottomrule
\end{tabular}
\end{table}

Here, \textit{PSA w/o sim} uses only the importance-based multi-level
mask $M$, while \textit{PSA w/ sim} additionally applies the
similarity-based cap $L$ from Alg.~\ref{alg:sim-constraint}.
Under the same sparsity budget, enabling the constraint improves the
overall Video-MME score from $0.647$ to $0.654$.

\subsection{Cosine Similarity (Video Generation)}
\label{sec:supp-sim-videogen}

We next investigate the cosine similarity constraint in the video generation
setting.
Table~\ref{tab:videogen-sim-ablation} compares three presets that share the same
PSA framework but differ in how multi-level pooling is assigned and how the
similarity thresholds are used.

\begin{table}[t]
\centering
\small
\setlength{\tabcolsep}{3pt}
\caption{Ablation on the cosine similarity constraint for video
generation (Wan2.1-1.3B, 480p). All presets operate at comparable
FLOP-equivalent sparsity.}
\label{tab:videogen-sim-ablation}
\begin{tabular}{lcccc}
\toprule
Preset        & Sparsity $\uparrow$ & PSNR $\uparrow$ & SSIM $\uparrow$ & LPIPS $\downarrow$ \\
\midrule
PSA (sim)     & 0.80 & 23.71 & 0.8615 & 0.1086 \\
PSA (no-sim)  & 0.79 & 23.36 & 0.8556 & 0.1186 \\
PSA (1-level) & 0.71 & 24.42 & 0.8787 & 0.0960 \\
\bottomrule
\end{tabular}
\end{table}

\textbf{PSA (sim).}
This preset adopts a more aggressive importance-based allocation in
order to maintain a FLOP-equivalent sparsity close to $0.8$ once the
similarity cap is applied.
Specifically, the thresholds $T$ in mask generation are set to $T=\{0.5,0.7,0.8,0.9\}$, together with similarity thresholds
$T_s=\{0.7, 0.65, 0.6\}$.
Since the cosine constraint may locally force heterogeneous KV blocks
to revert to finer levels, this more progressive allocation ensures
that the global sparsity remains in the target range while still
avoiding the assignment of coarse pooling to blocks whose neighboring tokens are not sufficiently similar.

\textbf{PSA (no-sim).}
To isolate the effect of the similarity constraint, we use a more
conservative importance-based assignment with thresholds
$T=\{0.7,0.8,0.9,0.9\}$, but disable the cosine cap by setting
$T_s=\{-1,-1,-1\}$.
This keeps a larger fraction of blocks at the finest level $h=1$ and
slightly reduces sparsity (0.79 vs.\ 0.80), yet all blocks are assigned
purely based on importance scores without checking whether a given KV
block is internally homogeneous enough to tolerate high-level pooling.

\textbf{PSA (1-level).}
Finally, \textit{PSA (1-level)} uses the same importance-based mask
ratios as \textit{PSA (no-sim)}, but sets all similarity thresholds to
$T_s=\{1,1,1\}$ so that the cosine cap always forces $L_j=1$ for any kept block.
As a result, the combined mask $\tilde{M}$ degenerates to a PSA with $H=1$, i.e., no pooling is applied and all kept blocks remain at the finest resolution.
This configuration therefore has the highest compute cost among the
three and serves as an upper bound on reconstruction quality for this
importance pattern.

Under comparable sparsity budgets, \textit{PSA (sim)} achieves a better
quality--sparsity trade-off than the purely importance-driven
\textit{PSA (no-sim)}, indicating that the cosine constraint helps
allocate coarse pooling more selectively: it avoids assigning high
pyramid levels to KV blocks whose tokens are not poolable, thereby
reducing approximation error at fixed sparsity.
Meanwhile, \textit{PSA (1-level)} confirms that further quality gains
are possible if we completely abandon pooling and pay a significantly
higher compute cost, providing a useful reference point for the
performance/efficiency envelope of PSA.

\begin{table}[t]
\centering
\small
\setlength{\tabcolsep}{4pt}
\caption{Ablation on block size and importance pooling for video
generation (Wan2.1-1.3B, 480p). All presets operate at an effective
sparsity of $0.8$.}
\label{tab:blocksize-pooling-ablation}
\begin{tabular}{ccccc}
\toprule
$(b_q, b_k)$ & Pool & PSNR $\uparrow$ & SSIM $\uparrow$ & LPIPS $\downarrow$ \\
\midrule
(128, 128) & Max  & 21.43 & 0.8475 & 0.1131 \\
(128, 128) & Mean & 21.08 & 0.8337 & 0.1257 \\
(32,  32)  & Max  & 21.64 & 0.8529 & 0.1102 \\
(32,  32)  & Mean & 21.71 & 0.8515 & 0.1101 \\
(64,  64)  & Max  & 21.55 & 0.8526 & 0.1099 \\
(64,  64)  & Mean & 21.56 & 0.8438 & 0.1179 \\
\bottomrule
\end{tabular}
\end{table}
\subsection{Block Size and Importance Pooling}
\label{sec:supp-ablation-pooling}

In this part, we ablate the logical block size $(b_q,b_k)$ and the pooling
operator used {inside the importance estimation step} under a fixed
sparsity budget.  
In PSA’s default formulation (\cref{sec:method-331}), the block
importance score is computed as
\[
S_{ij} = \max\!\left(\mathrm{Softmax}\!\left(
    \frac{\tilde{Q}_i \tilde{K}_j^\top}{\sqrt d}
\right)\right),
\]
i.e., using {max-pooling} over the sampled token-pair attention
weights.  
In Table~\ref{tab:blocksize-pooling-ablation}, we ablate this choice by
replacing the outer ``max'' with a {mean} operator:
\[
S_{ij}^{\text{mean}}
= \mathrm{mean}\!\left(\mathrm{Softmax}\!\left(
    \frac{\tilde{Q}_i \tilde{K}_j^\top}{\sqrt d}
\right)\right),
\]
while keeping every other component (sampling strategy, sparsity,
mask-generation rule, and multi-level allocation) unchanged.

At a fixed sparsity of $0.8$, PSA is relatively robust to the choice between
max- and mean-based importance pooling.
Max-pooling yields slightly higher PSNR/SSIM at $(128,128)$ and
$(64,64)$, while mean-pooling performs comparably at $(32,32)$.
Smaller block sizes—such as $(32,32)$—consistently yield better generation quality, as the finer partitioning more precisely isolates high-importance regions in the attention map, reducing approximation errors under the same sparsity budget.

\subsection{Multi-Level Allocation Strategy}
\label{sec:supp-multilevel}

Finally, we ablate the effect of the multi-level allocation strategy used by the quantile-based method. In this experiment, we fix all other hyper-parameters (block size, overall FLOP-equivalent sparsity, similarity constraint, and sampling strategy) and vary only how KV blocks are distributed across pyramid levels. All runs are conducted on Wan2.1-1.3B at 480p in the same training-free setting as above, with sparse attention enabled after the first \(25\%\) of sampling steps and the compute budget kept at approximately \(0.25\times\) full-attention FLOPs. We report the \emph{KV coverage ratio}, defined as the fraction of KV blocks that are retained (i.e., assigned to levels \(h>0\)); 

For clarity, we define five \emph{PSA presets}, denoted \textbf{PSA-1} to \textbf{PSA-5}, which all share the same compute budget but differ in how they allocate KV blocks across the pyramid levels. In this ablation we use three non-dropped levels \(h=1,2,3\) (with \(h=1\) the finest and \(h=3\) the coarsest), and a dropped level \(h=0\). The percentage of blocks assigned to each level, together with the resulting KV coverage, is summarized in Table~\ref{tab:psa-presets}.

\begin{table}[t]
\centering
\caption{
PSA presets: percentage of KV blocks per level.
L1/L2/L3 correspond to $h=1,2,3$; Drop corresponds to $h=0$.
}
\label{tab:psa-presets}
\setlength{\tabcolsep}{4pt} %
\begin{tabular}{c|cccc|c}
\toprule
\textbf{Preset} 
& \textbf{L1} 
& \textbf{L2} 
& \textbf{L3} 
& \textbf{Drop} 
& \textbf{Coverage} \\
\midrule
PSA-1 & 25\% & 0\%  & 0\%  & 75\% & 0.25 \\
PSA-2 & 0\%  & 0\%  & 100\% & 0\% & 1.00 \\
PSA-3 & 15\% & 10\% & 20\% & 55\% & 0.45 \\
PSA-4 & 10\% & 20\% & 20\% & 50\% & 0.50 \\
PSA-5 & 10\% & 10\% & 40\% & 40\% & 0.60 \\
\bottomrule
\end{tabular}
\end{table}

The quantitative results of this ablation are reported in Table~\ref{tab:multilevel-strategy-ablation}. All presets share the same logical block size \((b_q,b_k)=(128,64)\) and \emph{do not} enable the cosine similarity constraint, so any performance differences can be attributed solely to the multi-level allocation strategy.

\begin{table}[t]
\centering
\small
\setlength{\tabcolsep}{5pt}
\caption{Ablation on the multi-level allocation strategy for PSA on
Wan2.1--1.3B, 480p. We vary only how blocks are distributed across
pyramid levels (PSA-1 to PSA-5), while keeping the FLOP-equivalent
compute budget (about \(0.25\times\) full attention), block size, and
all other settings fixed.
``KV coverage'' denotes the fraction of KV blocks that are kept
(\(h>0\)).}
\label{tab:multilevel-strategy-ablation}
\begin{tabular}{lcccc}
\toprule
Preset & KV coverage $\uparrow$ & PSNR $\uparrow$ & SSIM $\uparrow$ & LPIPS $\downarrow$ \\ 
\midrule
PSA-1 & 0.25 & 20.94 & 0.8365 & 0.1286 \\
PSA-2 & 1.00 & 14.03 & 0.3447 & 0.8445 \\
PSA-3 & 0.45 & \textbf{22.16} & \textbf{0.8752} & \textbf{0.1004} \\
PSA-4 & 0.50 & 22.13 & 0.8683 & 0.1045 \\
PSA-5 & 0.60 & 21.94 & 0.8632 & 0.1070 \\
\bottomrule
\end{tabular}
\end{table}

Among these variants, PSA-2---which keeps {all} blocks but collapses them to a single coarse level to stay within the same FLOP budget---leads to severe degradation in all metrics. This shows that, under a fixed compute budget, concentrating almost all computation on one very coarse level yields poor reconstructions—even with maximal KV coverage—because the overly coarse KV representations dominate the performance degradation.

PSA-1, which preserves only a small portion of blocks at the finest representation and drops all others, performs worse than the multi-level strategies. This shows that allocating all compute to a few fine blocks, without using intermediate levels, severely restricts the effective receptive field for each query block and thus leads to degraded performance.

In contrast, the more balanced multi-level allocations in PSA-3/4/5
achieve consistently higher PSNR/SSIM and lower LPIPS at the same
FLOP-equivalent cost.
These strategies jointly trade off {where} to keep fine-grained
representations and {how much} area to cover with coarser
representations, instead of over-committing to a single level.
PSA-3 offers the best overall trade-off and is therefore adopted as our
default multi-level strategy in the main video generation experiments.
This ablation supports our design choice that, under a fixed compute
budget, extremely skewed mask distributions (either too many dropped
blocks or too many heavily pooled blocks) hurt quality, whereas a
balanced allocation across multiple pyramid levels yields superior
performance at similar sparsity.
Determining the \emph{optimal} allocation strategy, however, remains an
open question and is a promising direction for further exploration.

\section{Additional Visual Comparisons}
This section provides additional qualitative comparisons for \cref{traingfreevideogen}. 
Each figure follows the same layout as the examples in the main paper.

\begin{figure*}[t]
    \centering
    \includegraphics[width=\textwidth]{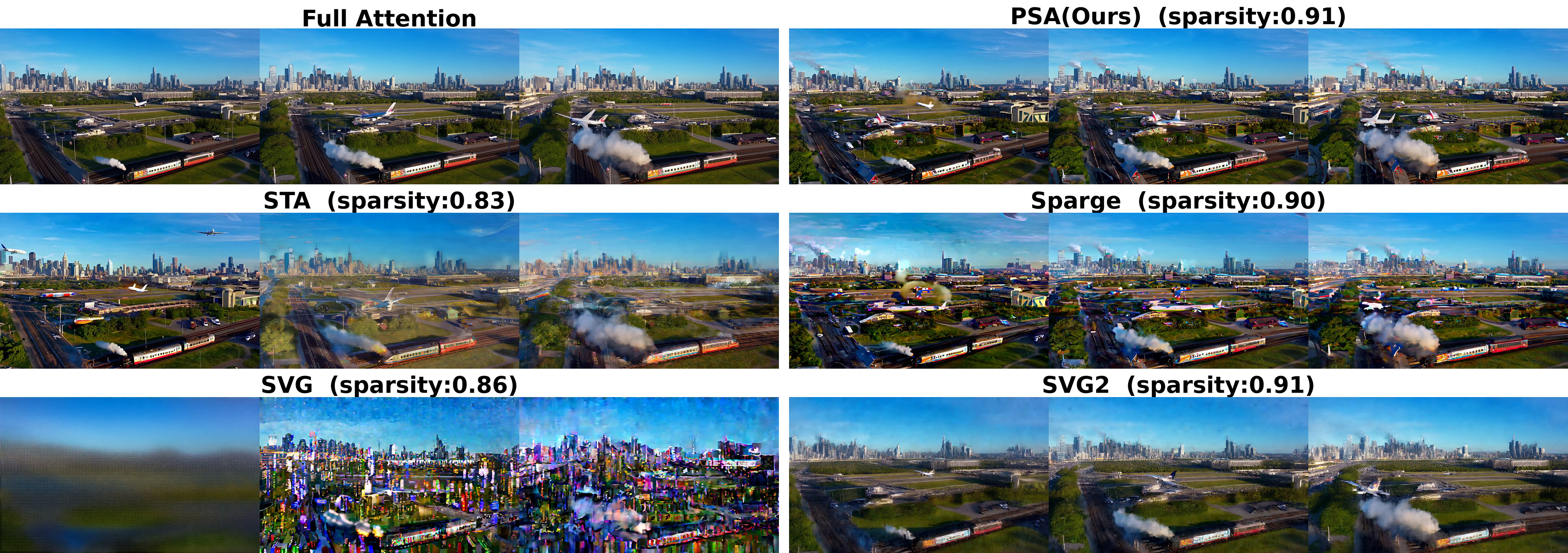}
    \caption{
A plane takes off over a city skyline; cut to a graffiti-covered steam train pulling into a station with billowing smoke. Cinematic colors and motion.
    }
    \label{fig:prompt001_comparison}
\end{figure*}

\begin{figure*}[t]
    \centering
    \includegraphics[width=\textwidth]{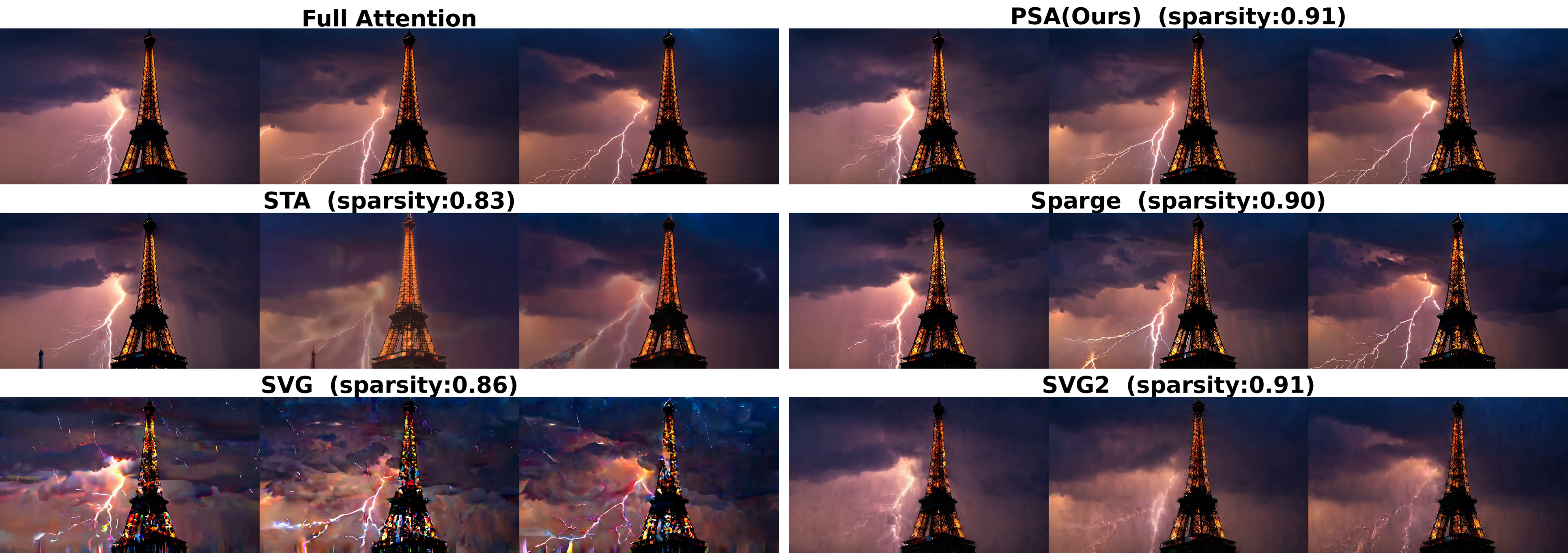}
    \caption{
A lightning bolt strikes the Eiffel Tower, illuminating its metal frame against swirling dark storm clouds. A low-angle view highlights the tower’s grandeur as the electric flash casts dramatic shadows.
    }
    \label{fig:prompt003_comparison}
\end{figure*}

\begin{figure*}[t]
    \centering
    \includegraphics[width=\textwidth]{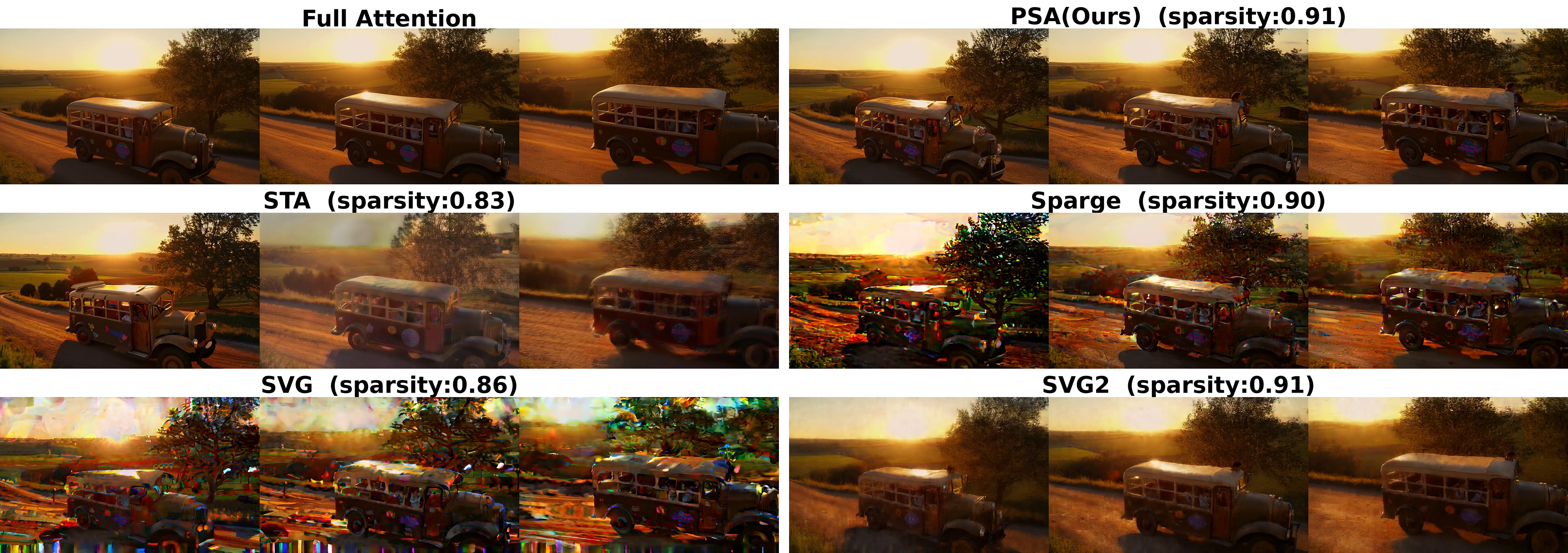}
    \caption{
A vintage school bus with retro decals turns a dusty rural corner at sunset. Warm golden light fills the scene as children inside read and play, and the focused driver steers through the tight bend. Low-angle, nostalgic cinematography with rolling hills in the background.
    }
    \label{fig:prompt004_comparison}
\end{figure*}

\begin{figure*}[t]
    \centering
    \includegraphics[width=\textwidth]{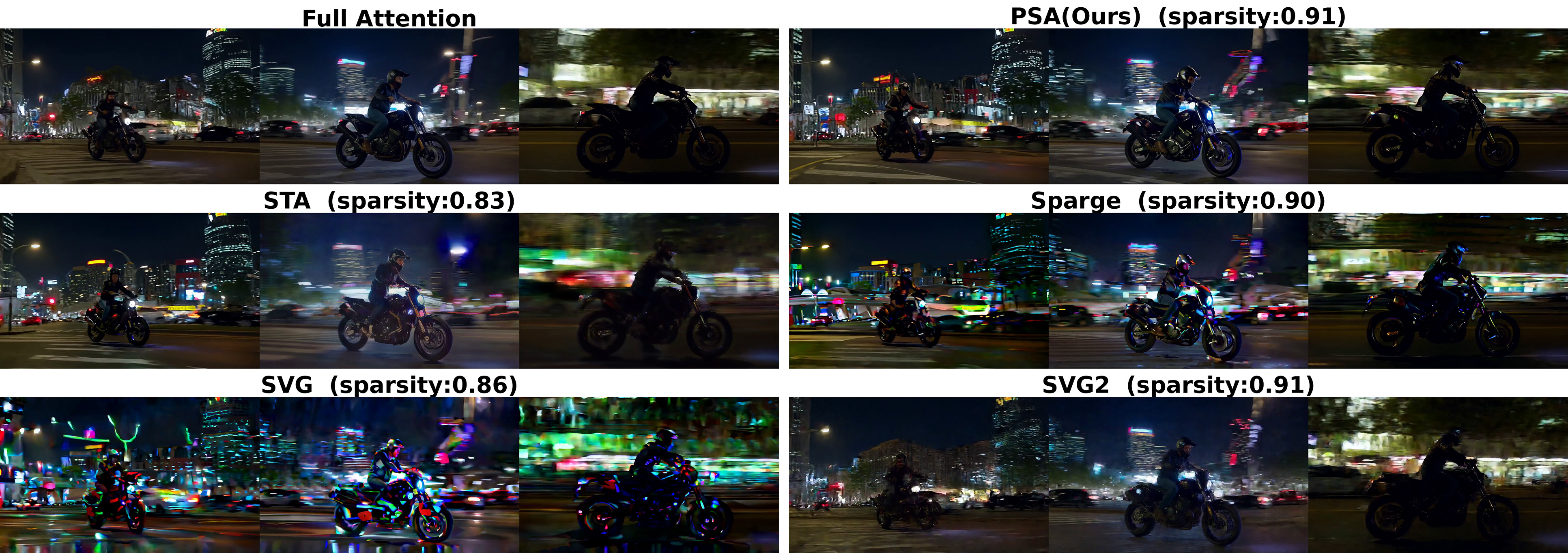}
    \caption{
A rider on a sleek black motorcycle weaves through neon-lit city streets at night, wearing a leather jacket and helmet. Dynamic shots follow their smooth maneuvers through traffic, with vibrant signs and skyscrapers creating an intense urban atmosphere.
    }
    \label{fig:prompt006_comparison}
\end{figure*}

\begin{figure*}[t]
    \centering
    \includegraphics[width=\textwidth]{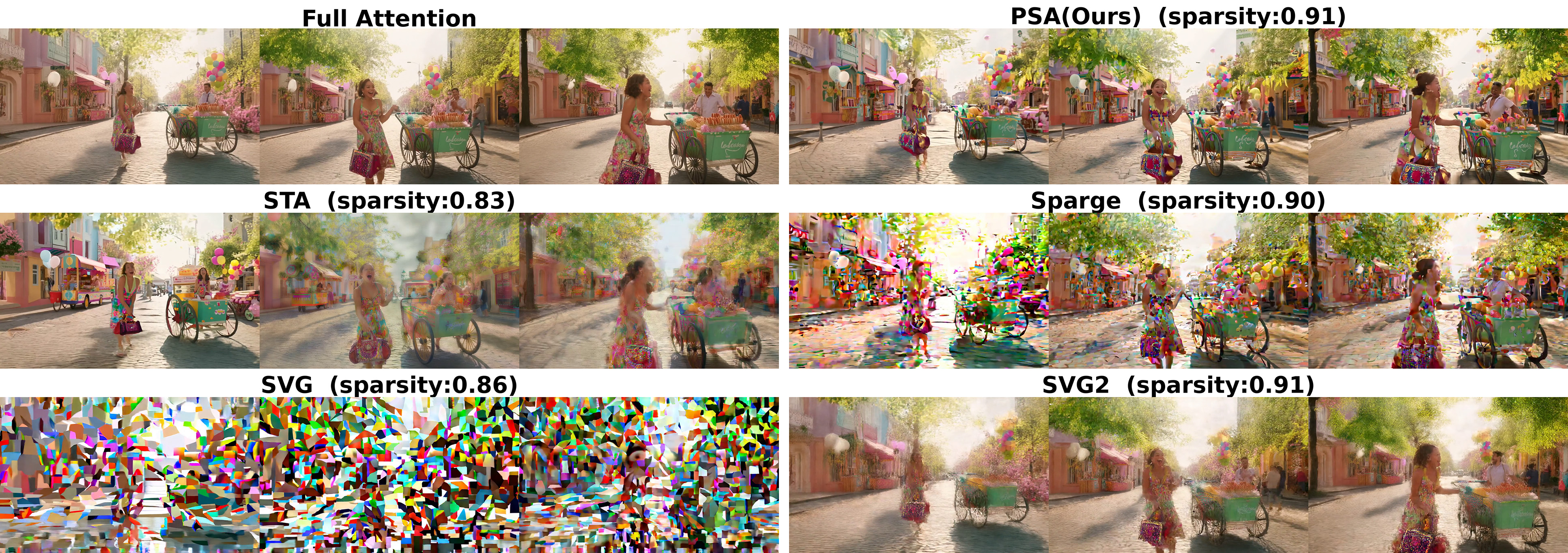}
    \caption{
A cheerful hot dog vendor pushes a colorful cart down a sunny pastel street as a smiling woman in a floral dress chats with passersby. Balloons, flowers, and lively vendors create a bright, carefree summer vibe.
    }
    \label{fig:prompt007_comparison}
\end{figure*}

\begin{figure*}[t]
    \centering
    \includegraphics[width=\textwidth]{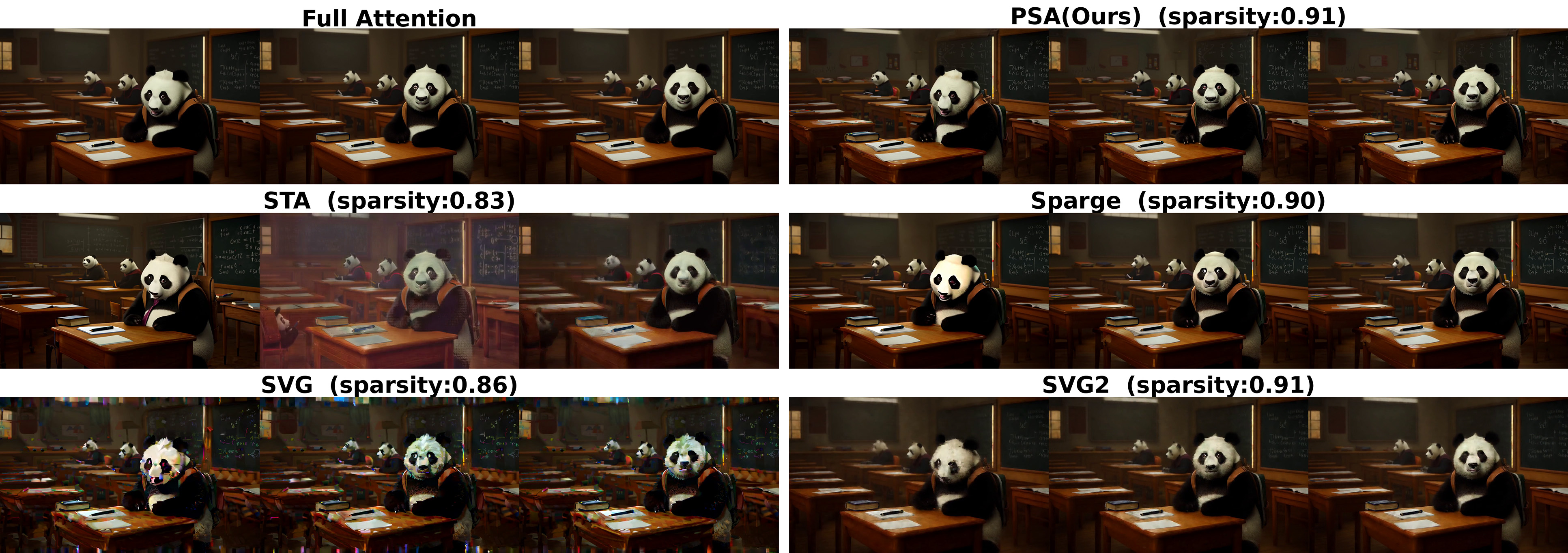}
    \caption{
    A puzzled panda student with a calculus book in a warm, dimly lit classroom, surrounded by desks, books, and attentive classmates.
    }
    \label{fig:prompt008_comparison}
\end{figure*}

\end{document}